\documentclass{article} % COLM 2026

\usepackage[preprint]{colm2026_conference}
\usepackage{microtype}
\usepackage{hyperref}
\usepackage{url}
\usepackage{booktabs}
\usepackage{lineno}
\usepackage{latexsym}
\usepackage{amsmath}
\usepackage{amssymb}
\usepackage{graphicx}
\usepackage{tikz}
\usetikzlibrary{positioning,arrows.meta,shapes.geometric,fit,backgrounds,calc}
\definecolor{darkblue}{rgb}{0,0,0.5}
\hypersetup{colorlinks=true, citecolor=darkblue, linkcolor=darkblue, urlcolor=darkblue}

\definecolor{nwmInk}{HTML}{1F2933}        % dark text/ink
\definecolor{nwmArrow}{HTML}{6B7280}      % mid-gray arrows
\definecolor{nwmTealFill}{HTML}{E3F1F0}   % pale teal (publish stage fill)
\definecolor{nwmTealLine}{HTML}{2C7A7B}   % dark teal (publish stroke/accent)
\definecolor{nwmTealBand}{HTML}{F2F8F8}   % very pale teal (publish band)
\definecolor{nwmAmberFill}{HTML}{FBF0DD}  % pale amber (retrieve stage fill)
\definecolor{nwmAmberLine}{HTML}{B7791F}  % dark amber (retrieve stroke/accent)
\definecolor{nwmAmberBand}{HTML}{FCF7EE}  % very pale amber (retrieve band)
\definecolor{nwmGrayFill}{HTML}{EEF1F4}   % light gray (store fill)
\definecolor{nwmGrayLine}{HTML}{94A3B8}   % medium gray (store stroke)
\definecolor{nwmGreen}{HTML}{2F855A}      % green accent / Answer
\definecolor{nwmGreenFill}{HTML}{E4F2EA}  % pale green (Answer chip)
\definecolor{nwmPurple}{HTML}{6B46C1}     % purple accent (temporal KG)
\definecolor{nwmPurpleFill}{HTML}{EDE7F8} % pale purple (KG node fill)
\definecolor{nwmIndigo}{HTML}{3C4FB1}     % indigo (reader)
\definecolor{nwmIndigoFill}{HTML}{E6E9F7} % pale indigo (reader fill)
\definecolor{nwmChipGray}{HTML}{D6DBE1}   % gray (Abstain chip)
\title{Narrative World Model: Narratology-Grounded Writer Memory for Long-Form Fiction}
\author{
Mohammad Saifullah, Thomas Kornmaier, Taaha Kazi,\\
\bf Vasu Sharma, Aditya Sanjiv Kanade, Aanand Kumar Yadav\\
\normalfont PocketFM\\
\normalfont\small\texttt{ullahsaif13407@gmail.com}\,\textrm{,}\,\texttt{mohammad.saifullah@pocketfm.com}
}
\date{}

\begin{document}

\ifcolmsubmission
\linenumbers
\fi

\maketitle

\begin{abstract}
Long-form fiction writers need memory that answers multi-hop questions about
evolving story state: who knows a secret and when they learned it, whether an
event preceded the narration that revealed it, whether a setup paid off, and how
a relationship shifted. General-purpose retrieval and agent-memory systems
represent entities and facts but not the narratological structure these
questions turn on, so they surface the wrong evidence or none at all. We
introduce the \emph{Narrative World Model} (NWM), a writer-memory system that
pairs a narratology-grounded typed temporal-state graph with query-conditioned
hybrid retrieval. To measure memory rather than the answerer, we read every
system through a single held-constant Opus~4.8 reader over only that system's
chapter-safe evidence, on a reproducible public corpus and a validated multi-hop
benchmark, and we compare against the strongest existing
temporal-knowledge-graph agent-memory framework, Graphiti/Zep
\citep{rasmussen2025zep}. NWM substantially and significantly outperforms this
baseline on multi-hop narratological QA across both corpora, and far exceeds
GraphRAG and flat retrieval. The advantage is representational rather than an
artifact of extraction: it survives rebuilding the baseline with NWM's own
extractor, and traces to its narratology-grounded structure and query-conditioned
retrieval, not to graph size or extractor quality.
\end{abstract}

\section{Introduction}

Long-form fiction generation exposes a gap between local fluency and durable
narrative state. A model can write an appealing scene while forgetting who knows
a secret, where an object is, whether a promise paid off, or how a relationship
changed. Chapter $N{+}1$ must respect finalized chapters $1\ldots N$ while
advancing the story.

\begin{figure}[t]
\centering
\resizebox{\textwidth}{!}{%
\begin{tikzpicture}[
    x=1mm, y=1mm,
    font=\footnotesize,
    >={Stealth[length=2.2mm]},
    every node/.style={text=nwmInk},
    stage/.style={
      rounded corners=1.8pt, draw, line width=0.7pt, fill=white,
      align=center, inner sep=3.5pt, minimum height=15mm, text width=27mm,
    },
    teal/.style={stage, fill=nwmTealFill, draw=nwmTealLine},
    amber/.style={stage, fill=nwmAmberFill, draw=nwmAmberLine},
    indigo/.style={stage, fill=nwmIndigoFill, draw=nwmIndigo},
    store/.style={
      cylinder, shape border rotate=90, aspect=0.22, draw, line width=0.7pt,
      fill=nwmGrayFill, draw=nwmGrayLine, align=center, inner sep=2.5pt,
      minimum width=34mm, minimum height=15mm, text width=28mm,
    },
    chip/.style={
      rounded corners=4pt, draw, line width=0.7pt, align=center,
      inner sep=2.5pt, minimum height=7mm, minimum width=20mm,
      font=\footnotesize\bfseries,
    },
    flow/.style={->, draw=nwmArrow, line width=0.9pt},
    fan/.style={->, draw=nwmArrow, line width=0.8pt},
    bandlabel/.style={font=\itshape\footnotesize, text=nwmArrow},
  ]

  % ================= Two-row layout (x,y in mm) =================
  % Row 1 (Publish, y=0): prose -> extraction -> two stores, side by side.
  % Row 2 (Retrieve & Read, y=-52): retrieval -> packet -> reader -> chips.
  % Stores feed DOWN into retrieval; writer query feeds UP into retrieval.

  % ---- Row 1: Publish phase (teal stages, neutral stores) ----
  \node[teal] (prose) at (0,0) {Finalized chapter prose\\[1pt]\scriptsize (C$_1$\ldots C$_n$)};
  \node[teal, text width=30mm, minimum height=22mm, right=10mm of prose] (extract)
    {Narratology-grounded extraction
     \\[2pt]{\scriptsize\raggedright
       $\bullet$~focalization / epistemic state\\
       $\bullet$~event- vs reveal-order\\
       $\bullet$~dramatic function\\
       $\bullet$~promise$\to$payoff\par}};

  \node[store, right=14mm of extract] (reg) {Typed current-state registries};
  \node[draw=nwmPurple, fill=nwmPurpleFill!55, rounded corners=2.5pt, line width=0.7pt,
        right=12mm of reg, minimum width=42mm, minimum height=20mm, inner sep=2pt] (kg) {};

  % green top accent on the registries cylinder
  \draw[line width=1.5pt, nwmGreen] ($(reg.north)+(-13mm,-1.4mm)$) -- ($(reg.north)+(13mm,-1.4mm)$);

  % Temporal KG: title at the top of the box, a clearly drawn small temporal graph below
  \node[font=\scriptsize, anchor=north] at ($(kg.north)+(0,-1.6mm)$) {Temporal knowledge graph};
  \node[circle, draw=nwmPurple, fill=white, line width=0.6pt, minimum size=3.4mm, inner sep=0pt]
        (kga) at ($(kg.center)+(-12mm,-1.4mm)$) {};
  \node[circle, draw=nwmPurple, fill=white, line width=0.6pt, minimum size=3.4mm, inner sep=0pt]
        (kgb) at ($(kg.center)+(0mm,-1.4mm)$) {};
  \node[circle, draw=nwmPurple, fill=white, line width=0.6pt, minimum size=3.4mm, inner sep=0pt]
        (kgc) at ($(kg.center)+(12mm,-1.4mm)$) {};
  \draw[->, nwmPurple, line width=0.7pt] (kga) -- (kgb);
  \draw[->, nwmPurple, line width=0.7pt] (kgb) -- (kgc);
  \node[font=\tiny, text=nwmPurple] at ($(kg.center)+(0,-6mm)$) {\textsf{valid\_at}\,/\,\textsf{invalid\_at}};

  % ---- Row 2: Retrieve \& Read phase (amber / indigo) ----
  \node[amber, minimum height=16mm] (retr) at (24,-52)
    {Query-conditioned hybrid retrieval
     \\[2pt]{\scriptsize BM25 + vector + 1-hop}};
  \node[amber, right=14mm of retr] (packet)
    {Chapter-safe evidence packet
     \\[1pt]{\scriptsize ($\leq$ chapter $n$)}};
  \node[indigo, right=14mm of packet] (reader)
    {Held-constant reader
     \\[1pt]{\scriptsize (Opus 4.8)}};

  % outcome chips, stacked to the right of the reader
  \node[chip, fill=nwmGreenFill, draw=nwmGreen,  text=nwmGreen, right=12mm of reader, yshift=6mm]
        (answer)  {Answer};
  \node[chip, fill=nwmChipGray,  draw=nwmGrayLine, text=nwmInk, right=12mm of reader, yshift=-6mm]
        (abstain) {Abstain};

  % ---- Pipeline arrows ----
  \draw[flow] (prose) -- (extract);
  \draw[fan] (extract.east) -- (reg.west);
  \draw[fan] (reg.east) -- (kg.west);
  % both typed stores feed retrieval (row 1 -> row 2) via one merged connector
  \coordinate (memrail) at ($(reg.south)!0.5!(kg.south)+(0,-9mm)$);
  \draw[fan,-,rounded corners=3pt] (reg.south) |- (memrail);
  \draw[fan,-,rounded corners=3pt] (kg.south)  |- (memrail);
  \draw[fan] (memrail) to[out=180,in=78] (retr.north);
  \draw[flow] (retr) -- (packet);
  \draw[flow] (packet) -- (reader);
  \draw[flow] (reader.east) -- (answer.west);
  \draw[flow] (reader.east) -- (abstain.west);

  % ---- Causal cutoff: dashed rule between the stores (row 1) and retrieval (row 2) ----
  \coordinate (cutY) at ($(reg.south)!0.5!(retr.north)$);
  \draw[dashed, nwmArrow, line width=0.7pt]
        ($(prose.west |- cutY)+(-2,0)$) -- ($(reader.east |- cutY)+(2,0)$);
  \node[font=\scriptsize, text=nwmArrow, fill=white, inner sep=1.5pt, anchor=center]
        at ($(packet.center |- cutY)$) {causal cutoff: chapters $>n$ hidden};

  % ---- Feedback: writer query feeds retrieval from below ----
  \node[stage, draw=nwmAmberLine, fill=white, text width=30mm, minimum height=8mm,
        below=11mm of retr] (query) {writer query for chapter $n{+}1$};
  \draw[->, dashed, draw=nwmAmberLine, line width=0.9pt] (query) -- (retr.south);

  \begin{scope}[on background layer]
    % Publish band: prose, extraction, both stores
    \node[rounded corners=5pt, fill=nwmTealBand, draw=nwmTealLine!35, line width=0.5pt,
          fit=(prose)(extract)(reg)(kg), inner ysep=6.5mm, inner xsep=5mm] (pubband) {};
    % Retrieve \& Read band: retrieval, packet, reader, outcome chips, query
    \node[rounded corners=5pt, fill=nwmAmberBand, draw=nwmAmberLine!35, line width=0.5pt,
          fit=(retr)(packet)(reader)(answer)(abstain)(query), inner ysep=6mm, inner xsep=5mm] (retband) {};
    \node[bandlabel, anchor=north west] at ($(pubband.north west)+(2mm,-2mm)$) {Publish};
    \node[bandlabel, anchor=south west] at ($(retband.south west)+(2mm,2mm)$) {Retrieve \& Read};
  \end{scope}
\end{tikzpicture}%
}
\caption{NWM pipeline. Finalized chapters are extracted into
narratology-grounded typed records (focalization/epistemic state,
event-vs-reveal order, dramatic function, promise/payoff) and a temporal
knowledge graph with validity intervals; query-conditioned hybrid retrieval
(BM25, vector, and one-hop graph expansion) assembles a chapter-safe evidence
packet from chapters up to the checkpoint only, and a held-constant reader
answers or abstains.}
\label{fig:nwm_architecture}
\end{figure}
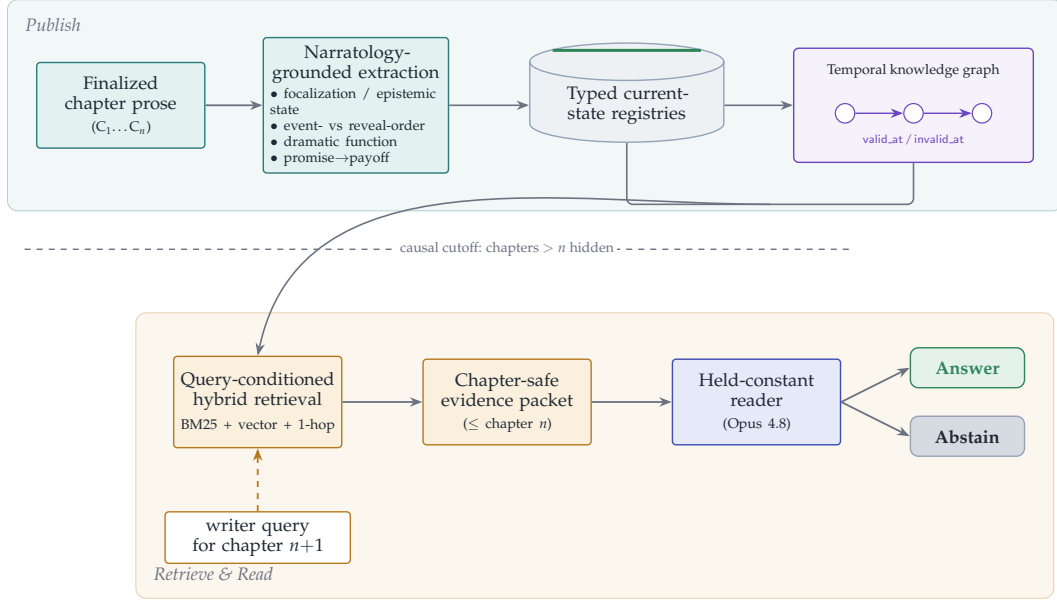

A writer-facing narrative memory system must do more than surface nearby text.
Rolling summaries discard details that later become plot-critical. RAG
\citep{lewis2020rag,guu2020realm} retrieves relevant prose, but a passage may say
where an object \emph{was} before it moved. Graph retrieval can expose entities
and events, but generic graphs do not necessarily track who knows what, the
difference between event order and reveal order, relationship deltas, unresolved
promises, or dramatic function. These approaches can miss the current, typed
narrative state a writer needs even when the source prose exists.

These failures concentrate on \emph{multi-hop narratological} queries: who knew
a fact at chapter $n$ and when they learned it, whether an event preceded the
narration that revealed it, whether a setup planted earlier paid off, and what
dramatic function a beat serves. Answering such a query requires evidence from
two or more distinct chapters and a representation that types narratological
structure rather than generic entities.

We therefore ask: does a narratology-grounded memory system provide better
chapter-safe evidence for multi-hop story-state questions than generic search,
source-chunk RAG, GraphRAG, or the strongest existing temporal-knowledge-graph
agent-memory framework, Graphiti/Zep \citep{rasmussen2025zep}? We
propose \emph{Narrative World Model} (NWM), which publishes finalized chapters
into typed memory records, global registries, a temporal knowledge graph (KG),
and a recursive language-model QA (RLM~QA) verifier. In a scoring protocol over
two corpora, each system answers nuanced writer questions from only its own
chapter-filtered memory or retrieval evidence, isolating memory/retrieval
behavior from generator behavior.

This paper makes the following contributions.
\begin{enumerate}
  \item A writer-facing multi-hop narrative-memory QA protocol and a validated
  176-item multi-hop slice built by curation, a single-passage hardness filter
  that discards questions a top-1 retrieved chunk can answer, and independent
  adjudication by a stronger model than the generator.
  \item The NWM system: a narratology-grounded typed temporal-state graph with
  query-conditioned hybrid retrieval, with its publish/update flow, schema
  registries, temporal KG, and RLM~QA.
  \item NWM Graph Retrieval significantly beats Graphiti, the strongest existing
  temporal-KG framework, on two corpora under a held-constant Opus~4.8 reader:
  $0.898$ versus $0.574$ on the private multi-hop slice (paired McNemar
  $64$--$7$, $p<10^{-5}$) and $0.625$ versus $0.516$ on the public 576-item set
  ($p=0.0001$), both far above GraphRAG and RAG.
  \item An extractor-fairness control showing the gain is representational:
  Graphiti is extracted with NWM's own Sonnet~4.5 model, and re-ingesting it with
  a cheaper extractor does not change its accuracy ($p=0.89$). A cross-system
  abstention analysis is the operative ablation.
  \item A reproducible public protocol and benchmark, including a 110-item
  public multi-hop subset.
\end{enumerate}

\section{Related Work}

\paragraph{Long-form story generation and revision.}
Neural story generation has long used intermediate representations to control
global structure. Hierarchical Neural Story Generation decomposes generation into
higher-level representations and lower-level realization
\citep{fan2018hierarchical}; Plan-and-Write uses an explicit storyline as a
control signal \citep{yao2019plan}; Re3 and DOC use recursive prompting,
revision, and detailed outlines to improve coherence
\citep{yang2022re3,yang2023doc}. These systems control \emph{intended} structure.
NWM addresses a different object: \emph{accepted} state after prose is finalized,
and whether that state is actually available to the next continuation.

\paragraph{Retrieval and long-term memory.}
RAG conditions generation on external evidence \citep{lewis2020rag}, and REALM
treats retrieval as latent memory \citep{guu2020realm}. Long-running memory
systems aim to carry information across interactions
\citep{zhong2023memorybank,wang2023longmem,packer2023memgpt,park2023generativeagents},
and production agent-memory frameworks such as Mem0 \citep{chhikara2025mem0}
extract and consolidate salient facts across sessions, with a graph variant that
captures relational structure. Even when a large context is available, models use
the middle unevenly \citep{liu2024lostmiddle}. These systems carry generic facts
or conversational state; NWM differs in its \emph{unit of state}: rather than
prose chunks or generic facts, it stores typed, versioned, chapter-scoped
narrative state and exposes the latest causally valid slice.

\paragraph{Knowledge graphs and narrative QA.}
Knowledge graphs represent relational facts for lookup and traversal
\citep{hogan2021knowledge}; GraphRAG-style systems build retrieval over
source-derived entity/event graphs \citep{edge2024graphrag}, and more recent
graph-structured retrievers such as HippoRAG \citep{gutierrez2024hipporag} and
LightRAG \citep{guo2024lightrag} couple entity graphs with dense retrieval to
support multi-document reasoning. For fiction the relevant graph is instead
temporal and narrative-specific. Multi-hop QA benchmarks such as HotpotQA
\citep{yang2018hotpotqa}, 2WikiMultiHopQA \citep{ho2020twowiki}, and MuSiQue
\citep{trivedi2022musique} establish that questions requiring evidence from two
or more passages form a distinct, harder regime than single-passage retrieval;
they motivate our multi-hop slice, though they target encyclopedic facts rather
than evolving narrative state. NarrativeQA established QA over stories as a test
of narrative understanding \citep{kocisky2018narrativeqa}. Our RLM~QA layer
follows the recursive-decomposition spirit of Recursive Language Models
\citep{zhang2025recursive} but specializes it to evidence-backed narrative-state
verification.

\paragraph{Narratology and temporal graph reasoning.}
Narratology distinguishes who sees from who speaks, story-event order from
discourse order, and event function within an arc
\citep{genette1980narrative,propp1968morphology}. Writer-facing craft
taxonomies also track dramatic situations and motivation/reaction beats
\citep{polti1921thirtysix,swain1965techniques}. Temporal-KG methods model
time-scoped facts and query-relevant graph neighborhoods
\citep{goel2020diachronic,han2021xerte}, while temporal KGQA benchmarks test
questions over time-scoped relations \citep{saxena2021cronquestions}. The
closest agent-memory framework is Graphiti/Zep \citep{rasmussen2025zep}, a
bi-temporal knowledge graph that records when a fact was true and when it was
ingested, and serves it as time-scoped evidence for agents. NWM is closest to
this line in its use of temporal edges and graph neighborhoods, but Graphiti's
edges are generic time-scoped facts, whereas NWM's schema types are
narrative-specific: knowledge boundaries, reveal order, promise/payoff status,
focalized observer, and dramatic function are first-class writer-memory fields
rather than generic entity relations. We use Graphiti as our strongest baseline.

\section{Narrative World Model}

\subsection{Causal Publish Flow}

NWM treats memory as a record of finalized prose, not future intent. Let
$C_n$ be accepted chapter prose, $M_n$ world state after chapter $n$, and
$O_{n+1}$ the next scaffold:
\begin{align}
M_n &= \mathrm{Publish}(C_n, M_{n-1}), \\
X_{n+1} &= \mathrm{Retrieve}(M_n, O_{n+1}),
\end{align}
where $X_{n+1}$ is the retrieved memory packet for chapter $n{+}1$ writer
queries and continuation support. The causal constraint is strict: a query or
continuation may use updates completed through chapter $n$, never future
scaffolds. Publishing stores a prose digest, extracts typed memory records, merges
registries, projects graph indexes, and exposes the completed state;
republishing an edited chapter invalidates downstream edges from that chapter.

$\mathrm{Publish}$ is an extraction step, not a copy: an extractor (Claude
Sonnet~4.5) reads the accepted prose and emits the typed records of §3.2
directly, each stamped with its source chapter and evidence span and merged into
cumulative registries (see Appendix~\ref{sec:method_details}).

\subsection{Schema Updates and Global Registries}

Instead of storing only summaries or prose chunks, NWM extracts a public
writer-memory structure. Implementations may add bookkeeping, but the
writer-facing retrieval state is this evidence-backed typed memory:

\begin{center}
\small
\textbf{NWM writer-memory structure}\\[2pt]
\begin{tabular}{p{0.28\linewidth}p{0.60\linewidth}}
\toprule
Memory record & Writer-visible fields \\
\midrule
Chapter digest & chapter id, source-grounded summary, evidence spans \\
Scene/event & location, participants, event order, reveal order, source chapter \\
Character state & goal, knowledge/unknowns, status, location, relationship deltas \\
Relationship state & character pair, relation type, polarity/status, validity \\
Object state & object, owner, location, condition, last evidence \\
Plot/promise & thread id, structural role, open/closed status, promised payoff \\
Narrative function & focalized observer, dramatic beat, turn/reversal, reader knowledge \\
World fact & fact, scope, valid chapter range, evidence \\
\bottomrule
\end{tabular}
\end{center}

Each record is chapter-scoped and evidence-backed, and merges into cumulative
registries (characters, relationships, locations, objects, active threads,
promises, world facts) that expose the latest causally valid slice for chapter
$n{+}1$. The writer query interface needs current slices for relevant entities,
not all prior chapters; thus a registry retains an old-but-current object
location that recent summaries omit.

Two properties make these records more than generic facts: every record is
\emph{evidence-backed} (it stores its source chapter and supporting span), and
the narratological fields are \emph{first-class typed slots} (focalized observer,
reveal order vs.\ event order, epistemic knowledge/unknowns, open/closed
promise-payoff status, dramatic function) rather than prose that happens to
mention a craft concept. These are the fields a generic entity/edge graph lacks
(see Appendix~\ref{sec:method_details}).

\subsection{Temporal Knowledge Graph}

The KG projects schema state into temporal relations: characters know facts,
objects are located at places, relationships change, threads resolve, and events
cause later events. Each edge stores source chapter, evidence, validity interval,
and confidence. A vector retriever returns a relevant mention; the graph returns
the \emph{latest valid} state as of chapter $n$ while preserving history.

Edges are typed by the narrative relation they assert (knowledge, location,
relationship-polarity change, thread resolution, causation), carrying the §3.2
semantics into the graph, and each edge's validity interval makes the graph
temporal rather than a flat snapshot. A query for state ``as of chapter $n$''
selects the latest edge valid at $n$ for each entity or relation while older
edges remain as history, returning an old-but-still-current state a rolling
summary cannot (see Appendix~\ref{sec:method_details}).

\subsection{Recursive Language-Model QA}

RLM~QA is a recursive verifier over NWM, not a new base model. It decomposes a
narrative question, queries schema and graph state, gathers evidence, and reports
a support trace. It assembles pre-continuation constraints and checks candidate
chapters for unsupported state changes.

\subsection{Query-Conditioned Hybrid Retrieval}

Retrieval turns the store into a chapter-safe evidence packet for a writer query
$q$ at checkpoint $n$: it restricts to records whose source chapter is $\leq n$,
ranks entity/seed nodes by a hybrid of BM25 \citep{robertson2009bm25} and dense
vectors \citep{xiao2023cpack} fused by reciprocal-rank fusion
\citep{cormack2009rrf}, expands each anchor's one-hop typed neighborhood, and
truncates to a bounded heterogeneous packet that is the only evidence the reader
sees (full four-stage description in Appendix~\ref{sec:retrieval}). This is what
distinguishes NWM Graph Retrieval from NWM State Memory, which serializes
\emph{all} state up to $n$ with no query-specific step: conditioning, not store
contents, is the operative difference between the two NWM conditions.

\subsection{Extraction and Baseline Independence}

The primary comparison deliberately separates baseline extraction from NWM
extraction. Simple search and source-chunk RAG index only chapter chunks;
GraphRAG builds its own entity/event graph directly from chapter text; Graphiti
\citep{rasmussen2025zep} ingests the same chapter episodes into its bi-temporal
knowledge graph and retrieves time-scoped facts; and the NWM rows use NWM's own
current-state memory and graph-retrieval conditions. No external baseline reads
NWM schemas, registries, graph state, or other NWM-derived memory records. Every
row is answered by the same held-constant Opus~4.8 answerer over its own
chapter-filtered evidence and evaluated with the same token-normalized support
rule; baseline construction itself may use an independent LLM extractor.

To separate the contribution of the representation from the contribution of the
extraction model, Graphiti is extracted with NWM's own model (Sonnet~4.5), so the
comparison is not confounded by extraction quality. As a robustness check, we also
re-ingest Graphiti with a cheaper extractor (GPT-4.1-mini); if a cheaper extractor
changed Graphiti's accuracy, the comparison would be sensitive to extraction
quality, whereas if it does not, the gap is representational. Both ingests are read
by the same Opus~4.8 answerer as every other row, holding the answerer and the
representation constant while varying only the extractor.

\section{Evaluation: Narrative Memory QA}
\label{sec:eval}

\subsection{Protocol}

We evaluate over two corpora. A public corpus of 12 public-domain books (six
genres, $\geq 20$ chapters each) makes the protocol reproducible. A private
corpus of five production-style serialized books of 50 chapters each is a
production-scale check; it is not redistributable.

External baselines use only chapter text and never consume NWM schemas,
registries, graph state, or other NWM-derived memory records:
\textbf{Simple Search} is lexical retrieval over chapter chunks;
\textbf{RAG} uses 360-word source chunks with 80-word overlap, BGE-large dense
embeddings \citep{xiao2023cpack}, BM25 \citep{robertson2009bm25}, and
reciprocal-rank fusion \citep{cormack2009rrf};
\textbf{GraphRAG} builds per-book source-derived entity/event graphs with an
independent LLM extractor and retrieves seed nodes plus local graph neighborhoods
per question; and
\textbf{Graphiti} \citep{rasmussen2025zep} ingests chapter episodes into a
bi-temporal knowledge graph and retrieves time-scoped facts, extracted with
Sonnet~4.5 to match NWM's extractor (a cheaper-extractor re-ingest is reported as
a robustness check in §\ref{sec:eval}).

The NWM rows evaluate two retrieval conditions over the same published memory:
direct current-state memory and question-conditioned graph retrieval. Every
system is filtered by book and chapter so it cannot read future chapters, and
every row is answered by the same held-constant Opus~4.8 answerer over its own
chapter-filtered evidence.

\paragraph{Private multi-hop benchmark.}
On the private corpus we curated a multi-hop benchmark: 176 validated multi-hop
questions, each requiring evidence from at least two distinct chapters, plus 96
matched single-hop control questions. Items are balanced across four
narratological families: focalization/epistemic (who knows what, and when they
learned it), reveal-order-versus-event-order, dramatic-shape/setup-payoff, and
combination. Questions were generated from chapter text with a strong language
model, then filtered by a single-passage hardness check that keeps a question
only if a top-1 retrieved 360-word chunk cannot support its answer, and finally
adjudicated for genuine multi-hopness and answer correctness by a separate,
stronger model than the generator. This curation isolates questions that demand
cross-chapter reasoning over typed narrative state.

\paragraph{Public multi-hop subset.}
On the public corpus we used the frozen 576-question source-written set and
defined a 110-question multi-hop subset as those questions whose gold evidence
cites at least two distinct chapters.

\paragraph{NWM retrieval conditions.}
\textbf{NWM State Memory} is the direct current-state condition: all typed memory
items published up to the checkpoint are merged and serialized into a bounded
evidence context, without an additional query-specific graph search step, testing
whether the published memory itself contains enough evidence. \textbf{NWM Graph
Retrieval} is the question-conditioned graph condition: the query ranks relevant
temporal-graph nodes and relations, expands local graph neighborhoods, and packages
compact evidence (plus recursive verification traces) for the same answerer,
combining structured current-state memory and query-conditioned graph retrieval
rather than a single vector index.

\paragraph{Answerers and scoring.}
All rows are answered by the same held-constant Opus~4.8 answerer over
\emph{only} the system's chapter-filtered evidence, so the reported differences
compare each system's memory representation, extraction, retrieval, and evidence
packaging under one answer model; the answerer must abstain when evidence is
insufficient and return evidence ids. Final correctness for all
rows is then computed with a token-normalized support check against the reference
answers, accepting direct answer-string support or overlap on answer-bearing
content tokens. Thus score measures evidence support for writer
questions, not generated-continuation behavior or privileged store access.
Because the same Anthropic family both adjudicates and reads the benchmark, we
replicate the held-constant reader with Google Gemini~3.1~Pro on the full
multi-hop set: the system ranking and the NWM-versus-Graphiti result hold
(Appendix~\ref{sec:crossfamily}).

\paragraph{Query taxonomy.}
Queries span latest character state, knowledge boundary, relationship constraint,
object location/state, event grounding, promise/payoff, temporal ordering,
world-rule constraint, narratological function, and source-needle detail. The
taxonomy is motivated by narratology and writing-craft categories such as
focalization, event function, dramatic turns, and promise/payoff, not by the
serialized NWM schema alone.

\paragraph{Public memory-QA comparison.}
The public 576-question set is curated from chapter text with GPT-5.5-assisted
curation; each item stores evidence spans, required chapters, checkpoint,
category, and curator provenance. The public corpus makes the protocol
reproducible while the private five-book corpus tests the same systems on longer,
production-style serialized stories; in both settings, accuracy gains require
useful evidence from the evaluated memory or retrieval system rather than from the
answerer alone.

\paragraph{Metrics.}
We report item-weighted QA accuracy under a deterministic token-coverage rule:
the gold answer's salient content tokens must be at least half-covered by the
evidence-supported answer. We report Wilson 95\% confidence intervals
\citep{wilson1927probable} and paired exact McNemar tests between systems on
identical items, which control for item difficulty and isolate where one
representation answers what another misses.

\section{Results}

All numbers below are produced by the evaluation implementation. The private
multi-hop comparison is the headline result; the public corpus replicates the
ordering on redistributable data. These are narrative-memory retrieval results,
not full-story continuation outcomes.

\subsection{Private Multi-Hop Comparison}

\begin{table}[t]
\centering\scriptsize
\setlength{\tabcolsep}{6pt}
\begin{tabular}{lccc}
\toprule
Memory source & Overall & Multi-hop & Control \\
\midrule
No Memory & 0.000 & 0.000 & 0.000 \\
Simple Search & 0.107 & 0.085 & 0.146 \\
RAG (BGE+BM25+RRF) & 0.184 & 0.176 & 0.198 \\
GraphRAG & 0.199 & 0.188 & 0.219 \\
NWM State Memory & 0.294 & 0.358 & 0.177 \\
Graphiti & 0.496 & 0.574 & 0.354 \\
NWM Graph Retrieval & \textbf{0.893} & \textbf{0.898} & \textbf{0.885} \\
\bottomrule
\end{tabular}
\caption{Private multi-hop comparison over 272 questions (176 validated
multi-hop, 96 single-hop control), scored with a held-constant Opus~4.8
answerer. Graphiti is extractor-matched to NWM (Sonnet~4.5). NWM Graph
Retrieval reaches $0.898$ multi-hop accuracy (Wilson 95\%
interval $[0.844, 0.934]$), against $0.574$ for Graphiti. On the 176 multi-hop
items the paired exact McNemar test between NWM Graph Retrieval and Graphiti is
$64$ to $7$, $p<10^{-5}$.}
\label{tab:private_multihop}
\end{table}

Table~\ref{tab:private_multihop} is the headline comparison. NWM Graph Retrieval
reaches $0.898$ on the 176 multi-hop items, against $0.574$ for the
extractor-matched Graphiti. The paired exact McNemar test between NWM Graph
Retrieval and Graphiti is $64$ to $7$, $p<10^{-5}$. Direct NWM State Memory, which
serializes current state without a query-conditioned graph search, reaches only
$0.358$ multi-hop and does not beat Graphiti; against NWM Graph
Retrieval the paired test is $101$ to $6$, $p<10^{-5}$. The advantage appears
only when the query ranks graph nodes and expands local neighborhoods. NWM Graph
Retrieval also leads on the single-hop control ($0.885$), showing the gain is not
specific to multi-hop items (per-system bars in
Figure~\ref{fig:multihop_bars}, Appendix~\ref{sec:method_details}).

\subsection{Public Comparison}

\begin{table}[t]
\centering\scriptsize
\setlength{\tabcolsep}{6pt}
\begin{tabular}{lcc}
\toprule
Memory source & Full (576) & Multi-hop (110) \\
\midrule
RAG & 0.314 & 0.291 \\
GraphRAG & 0.351 & 0.355 \\
NWM State Memory & 0.464 & 0.573 \\
Graphiti & 0.516 & 0.582 \\
NWM Graph Retrieval & \textbf{0.625} & \textbf{0.709} \\
\bottomrule
\end{tabular}
\caption{Public comparison over the frozen 576-question source-written set and
its 110-question multi-hop subset, scored with the same held-constant Opus~4.8
answerer. NWM Graph Retrieval reaches $0.625$ full-set and $0.709$ multi-hop
accuracy, above Graphiti at $0.516$ and $0.582$.}
\label{tab:public}
\end{table}

Table~\ref{tab:public} replicates the ordering on redistributable data. NWM
Graph Retrieval reaches $0.625$ on the full 576-item set and $0.709$ on the
110-item multi-hop subset, above Graphiti at $0.516$ and $0.582$ and far above
GraphRAG and RAG. The paired exact McNemar test between NWM Graph Retrieval and
Graphiti is $156$ to $93$ on the full set ($p=0.0001$) and $30$ to $16$ on the
multi-hop subset ($p=0.054$); the subset test is underpowered at $n=110$.

\paragraph{Extractor fairness.}
The advantage is representational, not an artifact of NWM's extractor. Graphiti is
extracted with NWM's own model (Sonnet~4.5), so the comparison is not confounded by
extraction quality. As a robustness check, re-ingesting Graphiti with a cheaper
extractor (GPT-4.1-mini) yields a sparser fact graph (Appendix~\ref{tab:system})---
entities rise from $443$ to $491$ while fact edges fall from $3{,}600$ to $2{,}226$
across the ingested chapter episodes---yet statistically indistinguishable accuracy
($0.585$ versus $0.574$; paired exact McNemar $p=0.89$), confirming the gap is
representational, not an extraction artifact. Matching or cheapening the extractor
does not close the gap to NWM Graph Retrieval.

\paragraph{Representation ablation.}
The cross-system comparison is itself the ablation. On the $64$ private
multi-hop items that NWM Graph Retrieval answered and Graphiti missed, Graphiti
abstained on all $64$, and in every case the gold fact was absent from
Graphiti's retrieved evidence. These were narratological-structure questions,
dramatic irony, reveal order, setup and payoff, and knowledge-boundary shifts,
for which Graphiti's generic entity/edge schema has no representation (per-family
case studies in Appendix~\ref{sec:family_case_studies}). The gap is
not extraction quality but the absence of narratological structure in the
representation.

Representation-size footprints (Table~\ref{tab:system},
Appendix~\ref{sec:method_details}) further show NWM's typed graph is smaller than
the source GraphRAG graph yet answers far more multi-hop questions: typing of
narrative structure, not graph density or extractor, is the driver
(per-family breakdown in Figure~\ref{fig:family}).

\section{Discussion}
\label{sec:discussion}

The results admit a single mechanistic reading: the advantage is where the
relevant narrative structure is \emph{represented and query-conditioned retrieval
surfaces it}, not where the graph is larger or the extractor is stronger.

\paragraph{Why structured narrative memory wins on multi-hop.}
A multi-hop narratological query is answered not by one fact but by composing
evidence about \emph{typed temporal units across chapters}: an epistemic boundary
that shifts, a reveal whose discourse position differs from its story position, a
promise opened and discharged. NWM decomposes chapters into exactly these
units---revelations, character/arc/object deltas, and scene beats, with validity
intervals---so query-conditioned retrieval assembles the cross-chapter structure
the query presupposes. A generic entity/edge graph
\citep{edge2024graphrag,rasmussen2025zep} has no such units. The failure mode
confirms this: Graphiti almost always \emph{abstained} rather than erring, with
the gold fact simply absent from its evidence---the signature of representational
absence, not reasoning error.

\paragraph{Retrieval conditioning is the operative mechanism.}
Structure in the store is necessary but not sufficient: the same typed memory
dumped as serialized current state (NWM State Memory) badly underperforms
query-conditioned NWM Graph Retrieval and does not even beat Graphiti. This is
mechanical, not a budget handicap: at an \emph{equal} evidence budget, State
Memory fills the window with a positional chapter prefix while query-conditioned
retrieval fills it with ranked, query-relevant records---so $83\%$ of its misses
are present-but-truncated and only $2\%$ absent (Appendix~\ref{sec:state_memory}).
Two controls bound the gain: an evidence oracle given the gold chapters answers
all $176$ items (a strict superset of NWM), so residual errors are retrieval- not
reader-limited, and dumping all prior chapters (${\sim}80$k tokens) does not beat
the retrieved slice (Appendix~\ref{sec:oracle}); and untyping NWM---stripping type
labels from the packet and re-embedding the graph so retrieval re-seeds---never
lowers accuracy (Appendix~\ref{sec:typing_ablation}).
The hop-count interaction reinforces this (Figure~\ref{fig:hopcount}); removing
the conditioning (State Memory) or representing the structure generically
(cross-system) collapses the advantage.

These results show NWM supplies the chapter-safe evidence a writer question
needs---a necessary precondition for NWM-conditioned \emph{generation}, the
writer-loop study we discuss in Section~\ref{sec:future}.

\section{Toward Generation and Writer-Workflow Evaluation}
\label{sec:future}

The next step is whether NWM-conditioned \emph{generation} is better. A
generation evaluation would place NWM inside the writing loop: continue story
worlds for $10$--$15$ chapters under matched memory conditions, then score
contradictions, entity/character/relationship consistency, harder composite QA,
writer correction burden, and blind pairwise human preference, testing whether
KG/RLM verification changes temporal disambiguation, contradiction detection, and
how retrieved memory shapes prose.

\section{Limitations}
\label{sec:limitations}

\paragraph{Isolating the typing.} The cross-system comparison shows a generic
schema (Graphiti) abstains on narratological-structure questions, but it varies
many factors at once. Two within-NWM controls isolate the labels: stripping the type labels from the
reader's packet, and re-embedding the graph untyped so retrieval re-seeds,
\emph{both} leave accuracy unchanged (Appendix~\ref{sec:typing_ablation}); packet
field-masking is uninformative because the reader recovers masked content from
surviving rows. The lever is thus the narrative decomposition and
query-conditioned retrieval, not the type labels. We do not isolate narratological
from generic fine-grained decomposition, nor confirm the retrieval-side null at
the production-vector level (its embed endpoint is access-gated); both are future
work.

\paragraph{Statistical power.} The public multi-hop subset is underpowered at
$n=110$, where the NWM-versus-Graphiti advantage ($30$ to $16$) is suggestive but
not significant ($p=0.054$). The private multi-hop slice ($n=176$, $p<10^{-5}$)
and the full public set ($n=576$, $p=0.0001$) carry the significant results.

\paragraph{Benchmark design.} The private multi-hop benchmark is
narratology-tilted by design, balanced across focalization/epistemic,
reveal-order, dramatic-shape, and combination families to stress exactly the
structure NWM types. This is the intended contribution, disclosed so the result
is read as a comparison on narratological multi-hop questions rather than
arbitrary story QA.

\paragraph{Automatic scoring.} The scorer is deterministic but not a
human-equivalence claim: a token-coverage rule can penalize paraphrases and accept
shallow matches, so we report evidence-support accuracy under a fixed rule; human
or LLM-judge agreement would strengthen it.

\paragraph{Data and workflow scope.} The public corpus is reproducible; the
five-book private corpus is production-style but not redistributable. All question
writing and external-baseline construction are separate from NWM's memory records.
The no-memory row is an evidence-free abstention control, not a closed-book recall
probe; conditioned \emph{generation} remains future work
(Section~\ref{sec:future}).

\paragraph{Baseline configuration.} GraphRAG is one concrete chapter-level
configuration (source-only extraction, per-book stores, seed-node retrieval, local
expansion), not Microsoft GraphRAG's global community-summary mode; Graphiti runs
in its default configuration, extractor-matched to NWM. The claim is a comparison
against these reproducible configurations, not every possible graph-RAG or
temporal-KG system.

\section{Conclusion}

Narratology-grounded writer memory significantly outperforms the strongest
existing temporal-knowledge-graph framework on multi-hop story-state QA. Under a
held-constant Opus~4.8 reader, NWM Graph Retrieval scored $0.898$ versus Graphiti
$0.574$ on a validated 176-item private multi-hop slice (paired McNemar $64$--$7$,
$p<10^{-5}$) and $0.625$ versus $0.516$ on the public 576-item set ($p=0.0001$),
far above GraphRAG and RAG. The gain is representational: Graphiti uses NWM's own
Sonnet~4.5 extractor, and re-ingesting it with a cheaper extractor leaves its
accuracy unchanged ($p=0.89$), and on
the multi-hop items NWM answers and Graphiti misses, Graphiti abstains on all
$64$ because its generic schema cannot represent narratological structure.
NWM-conditioned generation is future work.

\section*{Code and Data Availability}

The public benchmark is built from Project Gutenberg texts, and the evaluation
harness---question curation, the held-constant reader protocol, retrieval
baselines, and scoring---is reproducible on that corpus. The internal corpus,
its multi-hop benchmark, and the production Narrative World Model backend are
proprietary and are withheld; the private results are reported over anonymized
books (Book~A--E) with blinded titles and character names.

\bibliographystyle{colm2026_conference}
\bibliography{refs}

\begin{thebibliography}{34}
\providecommand{\natexlab}[1]{#1}
\providecommand{\url}[1]{\texttt{#1}}
\expandafter\ifx\csname urlstyle\endcsname\relax
  \providecommand{\doi}[1]{doi: #1}\else
  \providecommand{\doi}{doi: \begingroup \urlstyle{rm}\Url}\fi

\bibitem[Chhikara et~al.(2025)Chhikara, Khant, Aryan, Singh, and
  Yadav]{chhikara2025mem0}
Prateek Chhikara, Dev Khant, Saket Aryan, Taranjeet Singh, and Deshraj Yadav.
\newblock {Mem0}: Building production-ready {AI} agents with scalable long-term
  memory.
\newblock \emph{arXiv preprint arXiv:2504.19413}, 2025.

\bibitem[Cormack et~al.(2009)Cormack, Clarke, and Buettcher]{cormack2009rrf}
Gordon~V. Cormack, Charles L.~A. Clarke, and Stefan Buettcher.
\newblock Reciprocal rank fusion outperforms condorcet and individual rank
  learning methods.
\newblock In \emph{Proceedings of the 32nd International ACM SIGIR Conference
  on Research and Development in Information Retrieval}, 2009.

\bibitem[Edge et~al.(2024)Edge, Trinh, Cheng, Bradley, Chao, Mody, Truitt, and
  Larson]{edge2024graphrag}
Darren Edge, Ha~Trinh, Newman Cheng, Joshua Bradley, Alex Chao, Apurva Mody,
  Steven Truitt, and Jonathan Larson.
\newblock From local to global: A graph {RAG} approach to query-focused
  summarization.
\newblock \emph{arXiv preprint arXiv:2404.16130}, 2024.

\bibitem[Fan et~al.(2018)Fan, Lewis, and Dauphin]{fan2018hierarchical}
Angela Fan, Mike Lewis, and Yann Dauphin.
\newblock Hierarchical neural story generation.
\newblock In \emph{Proceedings of the 56th Annual Meeting of the Association
  for Computational Linguistics}, 2018.

\bibitem[Genette(1980)]{genette1980narrative}
G{\'e}rard Genette.
\newblock \emph{Narrative Discourse: An Essay in Method}.
\newblock Cornell University Press, 1980.

\bibitem[Goel et~al.(2020)Goel, Kazemi, Brubaker, and
  Poupart]{goel2020diachronic}
Rishab Goel, Seyed~Mehran Kazemi, Marcus Brubaker, and Pascal Poupart.
\newblock Diachronic embedding for temporal knowledge graph completion.
\newblock In \emph{Proceedings of the AAAI Conference on Artificial
  Intelligence}, 2020.

\bibitem[Guo et~al.(2024)Guo, Xia, Yu, Ao, and Huang]{guo2024lightrag}
Zirui Guo, Lianghao Xia, Yanhua Yu, Tu~Ao, and Chao Huang.
\newblock {LightRAG}: Simple and fast retrieval-augmented generation.
\newblock \emph{arXiv preprint arXiv:2410.05779}, 2024.

\bibitem[Guti{\'e}rrez et~al.(2024)Guti{\'e}rrez, Shu, Gu, Yasunaga, and
  Su]{gutierrez2024hipporag}
Bernal~Jim{\'e}nez Guti{\'e}rrez, Yiheng Shu, Yu~Gu, Michihiro Yasunaga, and
  Yu~Su.
\newblock {HippoRAG}: Neurobiologically inspired long-term memory for large
  language models.
\newblock In \emph{Advances in Neural Information Processing Systems}, 2024.

\bibitem[Guu et~al.(2020)Guu, Lee, Tung, Pasupat, and Chang]{guu2020realm}
Kelvin Guu, Kenton Lee, Zora Tung, Panupong Pasupat, and Ming-Wei Chang.
\newblock Realm: Retrieval-augmented language model pre-training.
\newblock In \emph{Proceedings of the 37th International Conference on Machine
  Learning}, 2020.

\bibitem[Han et~al.(2021)Han, Chen, Ma, and Tresp]{han2021xerte}
Zhen Han, Peng Chen, Yunpu Ma, and Volker Tresp.
\newblock Explainable subgraph reasoning for forecasting on temporal knowledge
  graphs.
\newblock In \emph{International Conference on Learning Representations}, 2021.

\bibitem[Ho et~al.(2020)Ho, Duong~Nguyen, Sugawara, and Aizawa]{ho2020twowiki}
Xanh Ho, Anh-Khoa Duong~Nguyen, Saku Sugawara, and Akiko Aizawa.
\newblock Constructing a multi-hop {QA} dataset for comprehensive evaluation of
  reasoning steps.
\newblock In \emph{Proceedings of the 28th International Conference on
  Computational Linguistics}, pp.\  6609--6625, 2020.

\bibitem[Hogan et~al.(2021)Hogan, Blomqvist, Cochez, d'Amato, de~Melo,
  Gutierrez, Kirrane, Gayo, Navigli, Neumaier, Ngomo, Polleres, Rashid, Rula,
  Schmelzeisen, Sequeda, Staab, and Zimmermann]{hogan2021knowledge}
Aidan Hogan, Eva Blomqvist, Michael Cochez, Claudia d'Amato, Gerard de~Melo,
  Claudio Gutierrez, Sabrina Kirrane, Jose Emilio~Labra Gayo, Roberto Navigli,
  Sebastian Neumaier, Axel-Cyrille~Ngonga Ngomo, Axel Polleres, Sabbir~M.
  Rashid, Anisa Rula, Lukas Schmelzeisen, Juan Sequeda, Steffen Staab, and
  Antoine Zimmermann.
\newblock Knowledge graphs.
\newblock \emph{ACM Computing Surveys}, 2021.

\bibitem[Hong et~al.(2025)Hong, Troynikov, and Huber]{hong2025contextrot}
Kelly Hong, Anton Troynikov, and Jeff Huber.
\newblock Context rot: How increasing input tokens impacts {LLM} performance.
\newblock Technical report, Chroma, July 2025.

\bibitem[Ko{\v{c}}isk{\'y} et~al.(2018)Ko{\v{c}}isk{\'y}, Schwarz, Blunsom,
  Dyer, Hermann, Melis, and Grefenstette]{kocisky2018narrativeqa}
Tom{\'a}{\v{s}} Ko{\v{c}}isk{\'y}, Jonathan Schwarz, Phil Blunsom, Chris Dyer,
  Karl~Moritz Hermann, G{\'a}bor Melis, and Edward Grefenstette.
\newblock The narrativeqa reading comprehension challenge.
\newblock \emph{Transactions of the Association for Computational Linguistics},
  2018.

\bibitem[Lewis et~al.(2020)Lewis, Perez, Piktus, Petroni, Karpukhin, Goyal,
  Kuttler, Lewis, Yih, Rocktaschel, Riedel, and Kiela]{lewis2020rag}
Patrick Lewis, Ethan Perez, Aleksandra Piktus, Fabio Petroni, Vladimir
  Karpukhin, Naman Goyal, Heinrich Kuttler, Mike Lewis, Wen-tau Yih, Tim
  Rocktaschel, Sebastian Riedel, and Douwe Kiela.
\newblock Retrieval-augmented generation for knowledge-intensive nlp tasks.
\newblock In \emph{Advances in Neural Information Processing Systems}, 2020.

\bibitem[Liu et~al.(2024)Liu, Lin, Hewitt, Paranjape, Bevilacqua, Petroni, and
  Liang]{liu2024lostmiddle}
Nelson~F. Liu, Kevin Lin, John Hewitt, Ashwin Paranjape, Michele Bevilacqua,
  Fabio Petroni, and Percy Liang.
\newblock Lost in the middle: How language models use long contexts.
\newblock \emph{Transactions of the Association for Computational Linguistics},
  2024.

\bibitem[Packer et~al.(2023)Packer, Fang, Patil, Lin, Wooders, and
  Gonzalez]{packer2023memgpt}
Charles Packer, Vivian Fang, Shishir~G. Patil, Kevin Lin, Sarah Wooders, and
  Joseph~E. Gonzalez.
\newblock {MemGPT}: Towards {LLMs} as operating systems.
\newblock \emph{arXiv preprint arXiv:2310.08560}, 2023.

\bibitem[Park et~al.(2023)Park, O'Brien, Cai, Morris, Liang, and
  Bernstein]{park2023generativeagents}
Joon~Sung Park, Joseph~C. O'Brien, Carrie~J. Cai, Meredith~Ringel Morris, Percy
  Liang, and Michael~S. Bernstein.
\newblock Generative agents: Interactive simulacra of human behavior.
\newblock In \emph{Proceedings of the 36th Annual ACM Symposium on User
  Interface Software and Technology}, 2023.

\bibitem[Polti(1921)]{polti1921thirtysix}
Georges Polti.
\newblock \emph{The Thirty-Six Dramatic Situations}.
\newblock J. K. Reeve, 1921.

\bibitem[Propp(1968)]{propp1968morphology}
Vladimir Propp.
\newblock \emph{Morphology of the Folktale}.
\newblock University of Texas Press, 1968.

\bibitem[Rasmussen et~al.(2025)Rasmussen, Paliychuk, Beauvais, Ryan, and
  Chalef]{rasmussen2025zep}
Preston Rasmussen, Pavlo Paliychuk, Travis Beauvais, Jack Ryan, and Daniel
  Chalef.
\newblock Zep: A temporal knowledge graph architecture for agent memory.
\newblock \emph{arXiv preprint arXiv:2501.13956}, 2025.

\bibitem[Robertson \& Zaragoza(2009)Robertson and Zaragoza]{robertson2009bm25}
Stephen Robertson and Hugo Zaragoza.
\newblock The probabilistic relevance framework: {BM25} and beyond.
\newblock \emph{Foundations and Trends in Information Retrieval}, 3\penalty0
  (4):\penalty0 333--389, 2009.

\bibitem[Saxena et~al.(2021)Saxena, Chakrabarti, and
  Talukdar]{saxena2021cronquestions}
Apoorv Saxena, Soumen Chakrabarti, and Partha Talukdar.
\newblock Question answering over temporal knowledge graphs.
\newblock In \emph{Proceedings of the 59th Annual Meeting of the Association
  for Computational Linguistics}, 2021.

\bibitem[Swain(1965)]{swain1965techniques}
Dwight~V. Swain.
\newblock \emph{Techniques of the Selling Writer}.
\newblock University of Oklahoma Press, 1965.

\bibitem[Trivedi et~al.(2022)Trivedi, Balasubramanian, Khot, and
  Sabharwal]{trivedi2022musique}
Harsh Trivedi, Niranjan Balasubramanian, Tushar Khot, and Ashish Sabharwal.
\newblock {MuSiQue}: Multihop questions via single-hop question composition.
\newblock \emph{Transactions of the Association for Computational Linguistics},
  10:\penalty0 539--554, 2022.

\bibitem[Wang et~al.(2023)Wang, Dong, Cheng, Liu, Yan, Gao, and
  Wei]{wang2023longmem}
Weizhi Wang, Li~Dong, Hao Cheng, Xiaodong Liu, Xifeng Yan, Jianfeng Gao, and
  Furu Wei.
\newblock Augmenting language models with long-term memory.
\newblock In \emph{Advances in Neural Information Processing Systems}, 2023.

\bibitem[Wilson(1927)]{wilson1927probable}
Edwin~B. Wilson.
\newblock Probable inference, the law of succession, and statistical inference.
\newblock \emph{Journal of the American Statistical Association}, 22\penalty0
  (158):\penalty0 209--212, 1927.

\bibitem[Xiao et~al.(2023)Xiao, Liu, Zhang, and Muennighoff]{xiao2023cpack}
Shitao Xiao, Zheng Liu, Peitian Zhang, and Niklas Muennighoff.
\newblock {C-Pack}: Packaged resources to advance general chinese embedding.
\newblock \emph{arXiv preprint arXiv:2309.07597}, 2023.

\bibitem[Yang et~al.(2022)Yang, Tian, Peng, and Klein]{yang2022re3}
Kevin Yang, Yuandong Tian, Nanyun Peng, and Dan Klein.
\newblock Re3: Generating longer stories with recursive reprompting and
  revision.
\newblock In \emph{Proceedings of the 2022 Conference on Empirical Methods in
  Natural Language Processing}, 2022.

\bibitem[Yang et~al.(2023)]{yang2023doc}
Kevin Yang et~al.
\newblock Doc: Improving long story coherence with detailed outline control.
\newblock In \emph{Proceedings of the 61st Annual Meeting of the Association
  for Computational Linguistics}, 2023.

\bibitem[Yang et~al.(2018)Yang, Qi, Zhang, Bengio, Cohen, Salakhutdinov, and
  Manning]{yang2018hotpotqa}
Zhilin Yang, Peng Qi, Saizheng Zhang, Yoshua Bengio, William~W. Cohen, Ruslan
  Salakhutdinov, and Christopher~D. Manning.
\newblock {HotpotQA}: A dataset for diverse, explainable multi-hop question
  answering.
\newblock In \emph{Proceedings of the 2018 Conference on Empirical Methods in
  Natural Language Processing}, pp.\  2369--2380, 2018.

\bibitem[Yao et~al.(2019)Yao, Peng, Weischedel, Knight, Zhao, and
  Yan]{yao2019plan}
Lili Yao, Nanyun Peng, Ralph Weischedel, Kevin Knight, Dongyan Zhao, and Rui
  Yan.
\newblock Plan-and-write: Towards better automatic storytelling.
\newblock In \emph{Proceedings of the AAAI Conference on Artificial
  Intelligence}, 2019.

\bibitem[Zhang et~al.(2025)Zhang, Kraska, and Khattab]{zhang2025recursive}
Alex~L. Zhang, Tim Kraska, and Omar Khattab.
\newblock Recursive language models.
\newblock \emph{arXiv preprint arXiv:2512.24601}, 2025.

\bibitem[Zhong et~al.(2023)Zhong, Guo, Gao, and Wang]{zhong2023memorybank}
Wanjun Zhong, Lianghong Guo, Qiqi Gao, and Yanlin Wang.
\newblock Memorybank: Enhancing large language models with long-term memory.
\newblock \emph{arXiv preprint arXiv:2305.10250}, 2023.

\end{thebibliography}

\appendix

\section{Additional Method Details}
\label{sec:method_details}

This appendix expands the method summaries in §3 and collects the supporting
figures and tables relocated from the body. The body retains the terse §3.1--3.4
prose and the schema table; the fuller descriptions below add no new claims.

\paragraph{Extraction at publish (expands §3.1).}
$\mathrm{Publish}$ is an extraction step, not a copy. When a chapter is
finalized, an extractor (Claude Sonnet~4.5) reads the accepted prose and emits
the typed memory records of §3.2 directly, rather than a free-text summary: each
emitted record is tied to the chapter that produced it and carries the evidence
span that licenses it. Records do not overwrite the store; they merge into the
cumulative registries, keyed by the entity, relationship, object, or thread they
describe, so that a single entity accretes a chapter-ordered history of typed
states. The causal cutoff is a property of the store, not of a single query: it
is enforced at write time, because every record is stamped with its source
chapter, and again at read time, because retrieval restricts to records stamped
at or before the checkpoint. Editing and republishing a chapter re-runs
extraction for that chapter and invalidates the downstream edges that depended on
the superseded state, so a correction propagates forward rather than leaving
stale assertions in the graph.

\paragraph{Evidence-backed narratological fields (expands §3.2).}
Two properties make these records more than generic facts, and both are visible
in the writer-memory structure (§3.2). First, every record is
\emph{evidence-backed}: it stores the source chapter it was extracted from and
the evidence span that supports it, so a downstream answer can cite where a state
was established rather than asserting it unsupported. Second, the narratological
fields are \emph{first-class typed slots}, not prose that happens to mention a
craft concept. A focalized observer names through whose perception a scene is
rendered; reveal order is stored separately from event order, so the chapter in
which the reader learns a fact is distinguished from the chapter in which the
underlying event occurred; character knowledge and unknowns record an epistemic
boundary, the set of facts a character does and does not yet hold at a
checkpoint; a plot or promise record carries an open-versus-closed payoff status,
marking whether a setup planted earlier has yet been discharged; and a narrative
function record names the dramatic beat or turn a scene performs. These are the
fields a generic entity/edge graph lacks: it can record that two entities are
related, but not who is licensed to know it, when it was revealed as opposed to
when it happened, or what arc function its disclosure serves.

\paragraph{Narrative-typed temporal edges (expands §3.3).}
Edges in the KG are typed by the narrative relation they assert rather than by a
generic association. An edge records that a character knows a fact, that an object
is located at or owned by a place or agent, that a relationship has changed
polarity, that a thread has resolved, or that one event causes a later event; the
same edge type therefore carries the narratological semantics of §3.2 into the
graph. Each edge stores its source chapter, its evidence, a validity interval
over chapters, and a confidence. The validity interval is what makes the graph
temporal rather than a flat snapshot: a fact that held from the chapter it was
established until the chapter it was superseded is a closed interval, while a fact
that still holds is left open.

\paragraph{State as of a checkpoint (expands §3.3).}
Because edges carry chapter-scoped validity, a query for state ``as of chapter
$n$'' is answered by selecting, among the edges valid at $n$, the latest one for
each entity or relation, while older edges remain in the graph as history rather
than being discarded. This is what a rolling summary cannot do: it lets the store
return an old-but-still-current state (for example, an object's last known
location) even when more recent chapters do not mention it, and it lets a later
query reconstruct the same entity's earlier state without re-reading the prose.

\subsection{Query-Conditioned Hybrid Retrieval (expands §3.5)}
\label{sec:retrieval}

The publish flow builds the store; retrieval is what turns it into a chapter-safe
evidence packet for a specific writer query. Given a writer query $q$ and a
checkpoint chapter $n$, $\mathrm{Retrieve}(M_n, q)$ proceeds in four stages.

\begin{itemize}
  \item \textbf{Causal restriction.} Retrieval first restricts the searchable
  store to records and edges whose source chapter is at most $n$, so no candidate
  evidence can originate in a future chapter. This is the same cutoff enforced at
  write time, re-applied as a hard filter at read time.
  \item \textbf{Hybrid node ranking.} Within that restricted store, entity nodes
  and seed nodes are ranked against $q$ by a hybrid score that combines a lexical
  signal (BM25 \citep{robertson2009bm25}) with a dense-vector signal over
  embedded node text \citep{xiao2023cpack}; the two rankings are merged by
  reciprocal-rank fusion \citep{cormack2009rrf} so neither signal dominates. The
  result is a small set of top-ranked entity and seed nodes that anchor the query
  in the graph.
  \item \textbf{One-hop graph expansion.} Each anchor node's one-hop neighborhood
  is expanded to pull in the typed edges incident to it (and their endpoints),
  bounded by a neighbor limit. This is the step that converts a ranked set of
  nodes into connected typed structure: it surfaces the knowledge, location,
  relationship-delta, reveal, and thread-resolution edges that a node-only ranking
  would leave implicit.
  \item \textbf{Bounded packet assembly.} The ranked vector hits, the expanded
  graph nodes, and the expanded graph edges are merged, re-ranked, and truncated
  to a size-bounded evidence packet. The packet is therefore heterogeneous by
  construction, ranked entity/seed vector hits together with the graph nodes and
  typed edges they induce, and it is the only evidence the held-constant reader
  sees for that query. RLM~QA (§3.4) may optionally decompose the query and verify
  the assembled evidence before the reader answers.
\end{itemize}

\paragraph{Why conditioning matters.}
The contrast with NWM State Memory makes the role of conditioning explicit. NWM
State Memory serializes \emph{all} typed memory items published up to $n$ into a
single bounded context, with no query-specific search step; it tests whether the
published state, taken whole, contains the answer. Query-conditioned hybrid
retrieval instead uses $q$ to decide \emph{which} parts of the store to surface
and then follows the graph one hop from those parts. The two share the same
underlying store and the same causal cutoff, and differ only in conditioning. A
multi-hop narratological query needs a few specific typed edges from different
chapters, an epistemic boundary established early, a reveal recorded later, a
relationship delta in between, and these are easy to bury when the entire state is
serialized at once but easy to surface when the query ranks the relevant nodes and
the graph supplies their typed neighborhoods. Conditioning, not store contents, is
therefore the operative difference between the two NWM conditions.

\subsection{Extended Discussion: Implications for Long-Form Generation}

These results establish that NWM supplies the chapter-safe evidence a writer
question needs; they do not by themselves establish that NWM-conditioned
\emph{generation} is better, a separate, necessary-but-not-sufficient step.
Answering correctly is a precondition for using the answer to constrain the next
chapter, but the writer-loop study (Section~\ref{sec:future}) is where one would
measure whether better evidence yields more consistent continuations. We expect
the regime where structured memory matters most to be the production regime of
very long serialized stories: as story length grows, full-context prompting
degrades both in cost and in the model's uneven use of long contexts
\citep{liu2024lostmiddle}, while a query-conditioned typed store returns a
bounded, chapter-safe slice whose size is set by the query rather than the story.
The margins here are on memory QA over corpora up to fifty chapters per book; the
gap to full-context prompting should, if anything, widen as stories lengthen
beyond what a single context can faithfully hold.

\begin{table}[t]
\centering\scriptsize
\resizebox{\linewidth}{!}{%
\begin{tabular}{lrr}
\toprule
Representation & Nodes / Entities & Edges \\
\midrule
NWM temporal graph (public) & 7{,}136 & 16{,}951 \\
Source GraphRAG graph (public) & 18{,}345 & 72{,}768 \\
Hybrid RAG source chunks (public) & 2{,}347 & --- \\
Graphiti (cheaper extractor) (private) & 491 & 2{,}226 \\
Graphiti (matched extractor) (private) & 443 & 3{,}600 \\
\bottomrule
\end{tabular}
}
\caption{Representation footprints from saved evaluation outputs; counts measure
representation size, not QA accuracy. On the 12-book public corpus the
source-only GraphRAG graph has $\sim 2.6\times$ as many nodes and
$\sim 4.3\times$ as many edges as the deployed NWM graph. On the private corpus,
Graphiti's matched (Sonnet~4.5) extractor yields a denser fact graph (more edges,
fewer entities) than the cheaper-extractor robustness ingest, yet the cheaper
ingest does not change its multi-hop accuracy.}
\label{tab:system}
\end{table}

Table~\ref{tab:system} quantifies representation size. NWM's typed graph is
smaller than the independently built source GraphRAG graph while answering far
more multi-hop questions (Tables~\ref{tab:private_multihop} and
\ref{tab:public}), and Graphiti's denser matched-extractor graph does not close the
gap. Read together, these two facts separate \emph{what} is represented from
\emph{how much}: if accuracy tracked graph size, the larger GraphRAG and denser
Graphiti graphs would lead, yet the smaller narratology-typed graph wins. The
result therefore points to narrative-structured decomposition and
query-conditioned retrieval, not graph density or extractor, as the driver.

\begin{figure}[t]
\centering
\includegraphics[width=\linewidth]{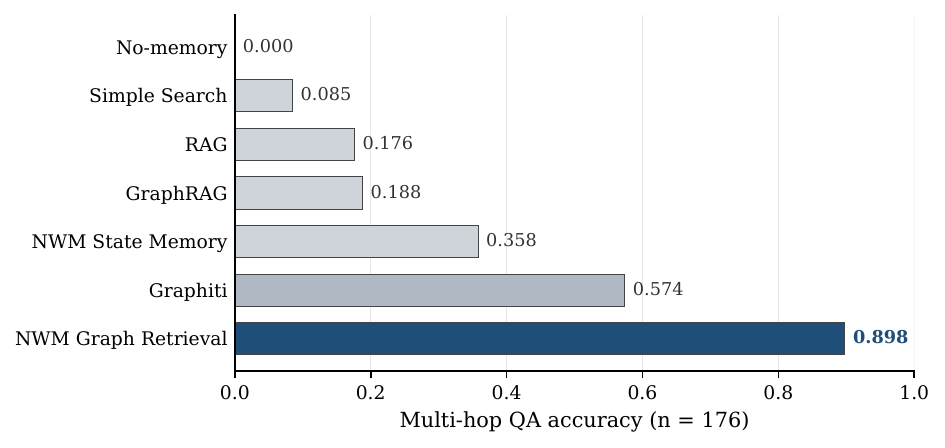}
\caption{Multi-hop accuracy for all systems on the private benchmark, plotting
the multi-hop column of Table~\ref{tab:private_multihop}.}
\label{fig:multihop_bars}
\end{figure}

\begin{figure}[t]
\centering
\includegraphics[width=\linewidth]{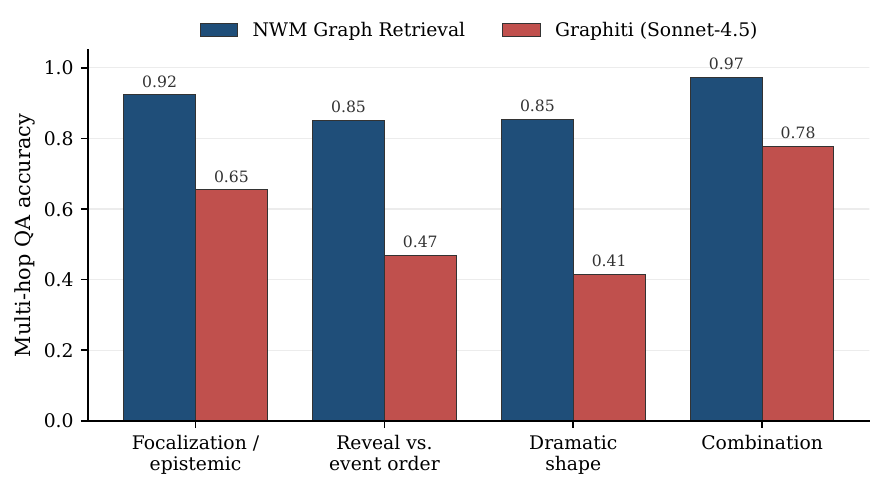}
\caption{Per-narratology-family multi-hop accuracy on the private benchmark, NWM
Graph Retrieval versus Graphiti (extractor-matched, Sonnet~4.5).}
\label{fig:family}
\end{figure}

\begin{figure}[t]
\centering
\includegraphics[width=\linewidth]{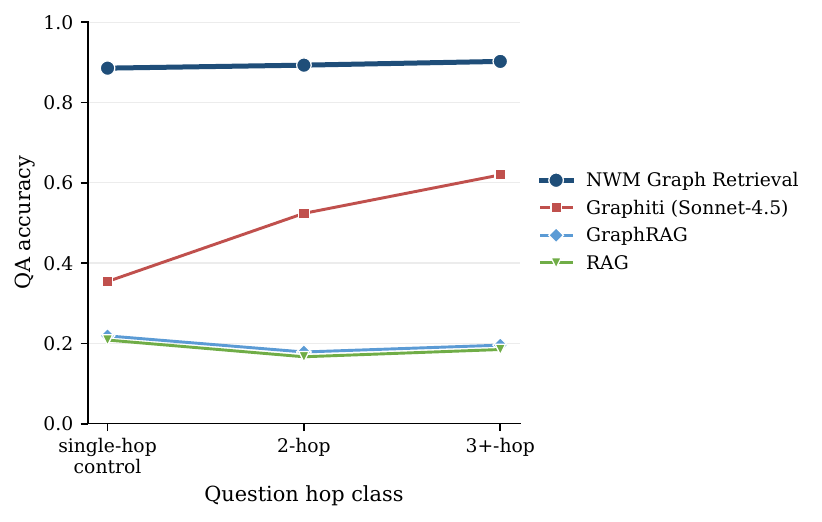}
\caption{Accuracy by hop class (single-hop control, 2-hop, 3+-hop).
Graph-structured systems rise or stay high as the hop count grows, while flat
retrievers stay low across hop classes.}
\label{fig:hopcount}
\end{figure}

\section{Memory-QA Examples}

Table~\ref{tab:qa_examples} shows twenty representative questions from the
public 576-item memory-QA set used in Table~\ref{tab:public}. Gold
answers are abbreviated for space; the released evaluation set contains the
full question, answer, evidence spans, curator id, book id, and checkpoint for
every item.

\begin{table}[t]
\centering\scriptsize
\setlength{\tabcolsep}{3pt}
\renewcommand{\arraystretch}{0.95}
\begin{tabular}{p{0.11\linewidth}p{0.08\linewidth}p{0.39\linewidth}p{0.34\linewidth}}
\toprule
Slice & Book & Example question & Gold answer \\
\midrule
Continuity &
pg9746 &
Why does Juliet Byrne not know who her birth parents are by the opening chapters? &
She was adopted by Lady Byrne and Sir Arthur, and Lady Byrne kept her parentage
secret. \\
Continuity &
pg2883 &
Who narrates the investigation in the opening chapters? &
Herbert Burroughs, a young but already experienced detective. \\
Continuity &
pg26514 &
Who is the narrator of the adventure? &
Mark Strong, owner of the yacht \emph{Celsis} and friend of Roderick and Mary. \\
Continuity &
pg25186 &
By checkpoint 5, who is secretly watching the Indian fleet and village? &
Henry Ware and his four forest-runner comrades are watching from cover. \\
Continuity &
pg1214 &
Why is Harmony left alone in the old lodge at the start? &
Scatch and the Big Soprano leave Vienna, while Harmony stays for music study. \\
\midrule
Narratology &
pg9746 &
How is Gimblet characterized when he first appears? &
A celebrated, observant private detective who loves strange and inexplicable
cases. \\
Narratology &
pg2883 &
How does the opening use Fleming Stone to define the detective ideal? &
Stone's brilliant deductions set a standard that Burroughs admires but cannot
match. \\
Narratology &
pg26514 &
How is Martin Hall characterized in the opening chapters? &
Comic and foolish on the surface, but his private request and manuscript reveal
courage. \\
Narratology &
pg25186 &
What mood does the opening river description establish? &
A vast, ancient, ominously quiet wilderness mood. \\
Narratology &
pg1214 &
What mood does the old stucco house create in chapter 1? &
Faded grandeur mixed with poverty, cold, and loneliness. \\
\midrule
Arc &
pg9746 &
What is the main inciting turn for Juliet by checkpoint 5? &
The solicitors' letter pulls Juliet from adopted life toward a family revelation. \\
Arc &
pg2883 &
What event launches Burroughs into the case? &
He is sent to West Sedgwick to investigate Joseph Crawford's murder. \\
Arc &
pg26514 &
What launches the main pursuit arc? &
Hall's request, disappearance, and manuscript push Mark toward Captain Black. \\
Arc &
pg25186 &
What arc function does the passing fleet serve in chapter 1? &
It launches the surveillance plot by leading the five to the Wyandot village. \\
Arc &
pg1214 &
What arc function does Scatchy's under-pillow gift serve? &
It forces Harmony's stay-or-return decision and makes independence precarious. \\
\midrule
Needle &
pg9746 &
What color was the envelope Lord Ashiel had Higgs bring to Gimblet? &
Blue. \\
Needle &
pg2883 &
What kind of bag is found on Crawford's desk? &
A gold-mesh woman's bag. \\
Needle &
pg26514 &
At what Paris hotel do Mark, Roderick, and Mary stay? &
The Hotel Scribe. \\
Needle &
pg25186 &
What crops are named in the cleared land around the Wyandot village? &
Corn and pumpkins. \\
Needle &
pg1214 &
What instrument does Harmony carry and study? &
The violin. \\
\bottomrule
\end{tabular}
\caption{Representative examples from the 576-question memory-QA set.}
\label{tab:qa_examples}
\end{table}

Table~\ref{tab:internal_qa_examples} shows representative questions from the
private five-book corpus used for the multi-hop benchmark in
Table~\ref{tab:private_multihop}. To preserve the confidentiality of the
production corpus, the five serialized books are de-identified as Books~A--E with
their genre, and character names are replaced by fixed neutral pseudonyms; the
questions and gold answers are otherwise written from chapter text rather than
NWM memory.

\begin{table}[t]
\centering\tiny
\setlength{\tabcolsep}{2pt}
\renewcommand{\arraystretch}{0.91}
\begin{tabular}{p{0.11\linewidth}p{0.115\linewidth}p{0.035\linewidth}p{0.37\linewidth}p{0.29\linewidth}}
\toprule
Slice & Book & Ch. & Representative internal question & Gold answer \\
\midrule
Continuity &
Book A (drama) &
10 &
At the chapter 10 checkpoint, what leverage does Vera use to draw Dana to the
Hotel Finest bar at 8pm? &
Vera claims she can tell Dana where her abandoned child is. \\
Continuity &
Book B (urban fantasy) &
20 &
After biting Lena, what secrecy problem does Reyes identify before she wakes up? &
He worries Lena may remember, and that exposure of his rare ability would draw
private companies and the military before he can protect himself. \\
Continuity &
Book C (dark fantasy) &
20 &
By chapter 20, why is Sable's True Name dangerous to reveal? &
Because the Shadow Bond would bind him to anyone who knows his True Name and
recites it aloud. \\
Continuity &
Book D (paranormal romance) &
20 &
After Lorne questions her about the forest, what does Selene conclude about
her hidden secret? &
Lorne has not discovered it; her secret and her life are still safe for now. \\
Continuity &
Book E (business drama) &
20 &
What investment deal does Ellison close with Hale for Nova TV in chapter 20? &
Ellison invests fifty million dollars for thirty percent of Nova TV's shares. \\
\midrule
Narratology &
Book A (drama) &
35 &
How does Mara's arrival reframe Dana's family standing in front of Royce's
household? &
Mara's poise immediately outclasses the Royce household, and Royce drops his
arrogance to ingratiate himself. \\
Narratology &
Book D (paranormal romance) &
10 &
What does the narration reveal during the dragon fight that Selene herself does
not notice? &
Her father arrives with Eclipse warriors, and another group arrives with a young
man who shifts into a massive golden dragon. \\
Narratology &
Book E (business drama) &
35 &
What reversal does Carver experience after Rourke explains Ellison's status in
chapter 35? &
Carver realizes Ellison is someone even Rourke cannot afford to offend, so he
becomes worried about insulting him. \\
\midrule
Arc &
Book A (drama) &
10 &
What separate 8pm plans are being set up at Hotel Finest in chapter 10? &
Aldo is staging a bar surprise while Cleo uses a walk with Mrs. Vane as
cover to meet Cassian at the first-floor cafe. \\
Arc &
Book B (urban fantasy) &
35 &
How does Reyes's blood bank change the Drall fight from a skill mismatch into a
winnable fight? &
It heals Reyes mid-fight, letting him take damage recklessly and keep pressing
Drall until the blood swipes break through. \\
Arc &
Book C (dark fantasy) &
50 &
Why do Nessa, Sable, and Cael decide they must attack the bone-scythe monster
before morning in chapter 50? &
They are trapped on the cliffs until morning; once sunrise lets the monster see
them, they lose surprise, so they choose to strike first. \\
Arc &
Book D (paranormal romance) &
50 &
What immediate political outcome has Selene engineered by chapter 50? &
The Alphas Summit has been cancelled, keeping Luna Coriel and Linus out of
danger. \\
\midrule
Needle &
Book A (drama) &
10 &
Where exactly does Cassian tell Cleo to meet him? &
At Table 28 in the cafe on the first floor. \\
Needle &
Book C (dark fantasy) &
50 &
What object saves Sable from being swept away by the rising sea during the storm? &
Nessa's golden rope. \\
Needle &
Book E (business drama) &
35 &
What does Alina realize Ellison's plain black card actually is in chapter 35? &
It is the legendary Apex Club black card, the club's highest-level membership
card, with only three in existence. \\
\bottomrule
\end{tabular}
\caption{Representative questions and gold answers from the 60-question internal
writer-memory QA set, with the five production books de-identified as Books~A--E
and character names replaced by fixed pseudonyms.}
\label{tab:internal_qa_examples}
\end{table}

\section{Private Multi-Hop QA Examples}
\label{sec:private_qa_examples}

Table~\ref{tab:private_qa_examples} shows eight representative items from the
176-question private multi-hop benchmark used in Table~\ref{tab:private_multihop},
two from each of the four narratological families. Each question requires
evidence from at least two distinct chapters (the \emph{Chapters} column lists
the gold required chapters). Book titles and character names are anonymized;
gold answers are abbreviated to roughly one line.

\begin{table}[t]
\centering\scriptsize
\setlength{\tabcolsep}{3pt}
\renewcommand{\arraystretch}{1.0}
\begin{tabular}{p{0.11\linewidth}p{0.11\linewidth}p{0.08\linewidth}p{0.31\linewidth}p{0.27\linewidth}}
\toprule
Family & Book & Chapters & Question & Gold answer (abbreviated) \\
\midrule
Focalization / epistemic &
Book A (drama) &
26, 38 &
How does a side character's knowledge of the twin secret evolve from confusion to partial understanding? &
At first he is baffled because the child playing with him and the child doing
homework both seem to be the same boy; after seeing both children he understands
that the protagonist is the boy's biological mother and why the twins keep the
truth secret for now. \\
\addlinespace
Focalization / epistemic &
Book B (progression fantasy) &
11, 12 &
How does the protagonist keep a peer from confirming a suspicion the peer already
holds about the protagonist's hidden ability? &
The peer already suspects a hidden ability after seeing the protagonist throw an
opponent with surprising ease; when the protagonist feels his wounds healing he
insists the peer leave, since watching the wound close would confirm it. \\
\midrule
Reveal vs.\ event order &
Book A (drama) &
1, 11 &
How does an antagonist's chapter-11 confession reframe the earlier explanation of
the protagonist's childhood obesity? &
Chapter 1 first explains the obesity as an accidental incorrect hormone
prescription, but the antagonist later reveals it was deliberate sabotage meant
to weaken the protagonist's marriage prospects. \\
\addlinespace
Reveal vs.\ event order &
Book B (progression fantasy) &
1, 18 &
How is the nature of the protagonist's inherited book withheld at first and then
reinterpreted by chapter 18? &
In chapter 1 the system's wording is cut off when the protagonist passes out;
only by chapter 18 does he reinterpret the accumulated signs and ask whether he
has been transformed into a supernatural being. \\
\midrule
Dramatic shape &
Book E (drama) &
1, 5 &
How does the story reverse a love interest's chapter-1 dismissal of the
protagonist by chapter 5? &
In chapter 1 she rejects him as too poor; in chapter 5 it is publicly revealed
that he secretly bought the high-end restaurant, leaving her shocked and unable
to grasp what she misjudged. \\
\addlinespace
Dramatic shape &
Book D (romance\slash fantasy) &
3, 9 &
What reversal occurs in the governess's narrative function between Chapters 3 and 9? &
In Chapter 3 the governess is the protagonist's abusive tormentor; by Chapter 9
she becomes a desperate mother begging the protagonist---the girl she
despises---for help saving her son, and the protagonist answers with sympathy
rather than gratification. \\
\midrule
Combination &
Book B (progression fantasy) &
7, 8 &
How do chapters 7 and 8 turn a roommate's ``fate'' claim into dramatic irony? &
Chapter 7 marks the roommate with a power-level-5 wrist number; in chapter 8 he
calls the shared room assignment ``fate,'' but a separate scene reveals someone
forced a student to swap rooms and that the attacker bore a 5, so the reader can
suspect engineered manipulation. \\
\addlinespace
Combination &
Book A (drama) &
2, 11, 12 &
How does a suitor's airport mistake set up his later humiliation at the bar? &
At the airport he fails to recognize the now-glamorous protagonist and denies any
tie to the ``ugly lady''; later he courts the same woman under an alias, only for
her to reveal it is merely her middle name and that she is the very person he
scorned. \\
\bottomrule
\end{tabular}
\caption{Representative private multi-hop QA items, two per narratological family.
Each requires evidence from the listed chapters. Titles and names anonymized;
gold answers abbreviated.}
\label{tab:private_qa_examples}
\end{table}

\section{Qualitative Case Study}
\label{sec:case_study}

We illustrate the representation gap with a single discordant item from the
\emph{dramatic-shape} family in Book D (romance/fantasy), spanning chapters 3 and
9, on which NWM Graph Retrieval answered correctly and Graphiti
(extractor-matched, Sonnet~4.5) abstained. This item is one of the 64
private multi-hop questions in the NWM-wins cell of the paired McNemar test
against Graphiti; Graphiti abstained on every one of those 64.

\paragraph{Question.}
\emph{What reversal occurs in the governess's narrative function between Chapters
3 and 9?}

\paragraph{Gold answer.}
In Chapter 3 the governess functions as the protagonist's abusive tormentor,
threatening to beat her and wishing she had drowned her. By Chapter 9 she becomes
a desperate mother begging the protagonist---the girl she despises---for help
saving her son, and the protagonist responds with sympathy rather than
gratification.

\paragraph{NWM Graph Retrieval (correct).}
Query-conditioned retrieval over the NWM temporal graph surfaced two
chapter-anchored scene-beat nodes that together encode the role reversal: an
early beat in which the governess is the powerful abuser, and a chapter-9 beat in
which she is an injured, helpless supplicant begging for help with her abducted
son. Reading only these chapter-safe rows, the held-constant answerer recovered
the dramatic-function shift: ``The governess shifts from being the protagonist's
mocking, powerful abuser to an injured, helpless supplicant who begs her for help
saving her abducted son, reversing their power dynamic.'' The narratological
link---a single character's \emph{dramatic function} inverting across chapters---
was carried by the graph's typed scene/situation structure, not by any one
surface fact.

\paragraph{Graphiti (abstained).}
Graphiti's retrieval returned a single entity/fact edge for the same query, and
the answerer responded ``Not enough evidence is available to answer from the
retrieved context'' and abstained. Graphiti's generic entity/relationship schema
records that the governess and the protagonist co-occur and interact, but has no
representation of a character's evolving \emph{narrative role}; the
chapter-3-versus-chapter-9 functional reversal was simply absent from its
retrieved evidence. This pattern is typical of the discordant cases: across the
64 multi-hop items NWM answered and Graphiti missed, Graphiti abstained on all 64,
and in every case the gold fact was missing from its retrieved evidence
altogether.

\section{Benchmark Construction Details}
\label{sec:benchmark_construction}

The private multi-hop benchmark was built over the five production-style
serialized books (50 chapters each) in three stages.

\paragraph{Curation.}
Candidate questions were generated from chapter text with a strong language
model (GPT-5.5, high-reasoning setting), instructed to target the four
narratological families---\allowbreak focalization\slash epistemic state, reveal order versus
event order, dramatic shape (setup and payoff), and combinations of these---\allowbreak and
to require evidence that spans more than one chapter.

\paragraph{Single-passage hardness filter.}
Each candidate was passed through a single-passage hardness check: the question
was retained only if a top-1 retrieved 360-word chunk \emph{could not} support
the answer. This removes questions that a flat passage retriever could solve from
one local window and biases the surviving set toward genuinely multi-chapter
reasoning.

\paragraph{Independent adjudication.}
The surviving questions were adjudicated by a separate, stronger model than the
generator (Opus 4.8) for two properties: genuine multi-hopness (the gold answer
demonstrably depends on evidence from at least two distinct chapters) and answer
correctness against the cited evidence spans. Items failing either check were
dropped or flagged.

\paragraph{Final set.}
The procedure yielded 176 validated multi-hop questions plus 96 matched
single-hop control questions, balanced across the four narratological families
(focalization/epistemic 52, reveal-vs.-event-order 47, dramatic-shape 41,
combination 36 among the multi-hop items). The control questions are single-chapter
by construction and are used to confirm that systems are not simply rewarded for
verbosity: a system must still abstain when the chapter-filtered evidence is
insufficient. Every released item carries its full question, gold answer, evidence
spans with chapter anchors, narratological family, hop count, and the curator and
adjudicator identifiers.

\section{State Memory: A Sufficient Store, Truncated}
\label{sec:state_memory}
NWM State Memory serializes the same typed store NWM Graph Retrieval draws on,
yet scores only $0.358$ multi-hop. The cause is delivery, not representation. The
serialized current-state context is large---median ${\sim}228$k characters
(${\sim}57$k tokens), chapter-ordered---while the held-constant reader applies
the \emph{same} evidence budget to every system ($12$k characters, ${\sim}3$k
tokens). All $176$ State Memory contexts therefore overflow and are truncated to
a positional prefix (roughly chapters~1--3), discarding ${\sim}95\%$ of the
store. On the discordant set (State Memory wrong, Graph Retrieval right;
$n{=}101$):

\begin{center}\small
\begin{tabular}{lcc}
\toprule
Where is the gold fact? & count & \% \\
\midrule
present, within the $12$k budget (reader miss) & 15 & 14.9 \\
present but past the truncation cut & 84 & 83.2 \\
absent from the store & 2 & 2.0 \\
\bottomrule
\end{tabular}
\end{center}

\noindent Only $2\%$ of misses are genuine representation gaps; $83\%$ are facts
the store contains but positional truncation drops. Query-conditioned retrieval
spends the identical budget on ranked, query-relevant records instead of a
chapter prefix, which is why the same store yields $0.898$ when queried and
$0.358$ when dumped. The mechanism is ranking under a fixed, shared budget---not
memory content, model, or a budget handicap.

\section{Evidence Oracle (Reader Upper Bound)}
\label{sec:oracle}
Given the full text of the gold cited chapters, the held-constant Opus~4.8 reader
answers all $176$ multi-hop items correctly ($1.000$; zero abstentions), on every
narratology family. To fit the gold chapters untruncated we raise the reader's
render budget to $60$k characters (median gold evidence ${\sim}14$k characters);
model, scorer, and decoding are otherwise identical to the main runs. The oracle
is a strict superset of NWM Graph Retrieval (paired exact McNemar: $0$ items
NWM-only, $18$ items oracle-only, $p\approx8\times10^{-6}$): it fixes every NWM
failure and breaks none. Two conclusions follow: the benchmark is answerable by
the reader when the evidence is present, and NWM's residual ${\sim}10$-point gap
is attributable to retrieval not surfacing the right chapters (concentrated in
reveal-order and dramatic-shape), not to reader capability.

\paragraph{Full-context (all prior chapters).} As a companion upper bound we give
the reader the full text of \emph{all} chapters up to the checkpoint (median
${\sim}80$k tokens) instead of the gold chapters or a retrieved slice.

\begin{center}\small
\begin{tabular}{lccc}
\toprule
Reader receives & Acc & vs.\ NWM (exact McNemar) & tokens \\
\midrule
gold chapters (oracle) & 1.000 & $b{=}18, c{=}0$, $p\approx8\times10^{-6}$ & ${\sim}14$k \\
NWM retrieved slice & 0.898 & --- & ${\sim}12$k \\
all prior chapters & 0.852 & $b{=}22, c{=}14$, $p{=}0.24$ & ${\sim}80$k \\
\bottomrule
\end{tabular}
\end{center}

\noindent Dumping every prior chapter is \emph{not} a substitute for retrieval:
all-prior-context is statistically indistinguishable from NWM's retrieved slice
($p{=}0.24$) despite ${\sim}7\times$ more tokens, and far short of the gold-only
oracle---the evidence is present but the reader cannot surface it (needle dilution
/ lost-in-the-middle), worst on the long-range dramatic-shape family.
Both upper bounds are reachable only because our chapters and gold evidence fit a
single context window. As serialized stories lengthen, the gold evidence itself
outgrows any reader's window, and even within it long-context quality decays
through lost-in-the-middle \citep{liu2024lostmiddle} and \emph{context rot}
\citep{hong2025contextrot}---so
the oracle itself degrades and full-context prompting becomes infeasible, exactly
where a bounded, query-sized retrieval like NWM's is necessary rather than merely
competitive. Quantifying this crossover as stories scale past the window is future
work.

\section{Typing Ablations: Reader-Side and Retrieval-Side}
\label{sec:typing_ablation}
To separate NWM's narratological \emph{type labels} from the \emph{content} its
retrieval surfaces, we hold retrieval fixed (identical hits, ranks, and scores
from the faithful prod-vector run) and strip the typed scaffolding the reader
sees: relation/type labels and typed \texttt{key=value} tags are removed or
collapsed to generic relations, while the descriptive snippet content is
preserved (a flattened variant additionally inlines the remaining typed metadata
as plain prose).

\begin{center}\small
\begin{tabular}{lcc}
\toprule
Condition (retrieval held fixed) & Multi-hop acc & McNemar $p$ vs.\ full \\
\midrule
NWM Graph Retrieval (full typed) & 0.898 & --- \\
type labels stripped (untyped) & 0.909 & 0.62 \\
typed records flattened to prose & 0.926 & 0.12 \\
\bottomrule
\end{tabular}
\end{center}

\noindent Removing the type labels does not lower accuracy (both variants are
statistically indistinguishable from the full packet and trend slightly higher,
plausibly because dropping verbose scaffolding frees budget for content). The
held-constant reader thus does not depend on reading the narratological type
labels: once query-conditioned retrieval has surfaced the right records, their
\emph{content} carries the answer. This rules out a reader-side packaging
artifact.

That is a reader-side test only---retrieval still used the fully typed graph. We
therefore also untype \emph{retrieval}: we collapse all relation types to a single
generic edge and strip the narratological type tokens from the node text used for
embedding, then \emph{re-embed and re-seed} so the retrieved set genuinely changes
(top-$10$ seed overlap falls from a Jaccard of $1.0$ to a mean of $0.31$, with
$0/176$ items keeping an identical seed set).

\begin{center}\small
\begin{tabular}{lcc}
\toprule
Condition (re-embedded, same pipeline) & Multi-hop acc & McNemar vs.\ typed \\
\midrule
typed graph & 0.807 & --- \\
untyped graph (re-seeded) & 0.886 & $p=0.009$ \\
\bottomrule
\end{tabular}
\end{center}

\noindent Untyping does not reduce accuracy here either; it slightly \emph{raises}
it ($p=0.009$), consistent with type-name tokens acting as embedding-space noise
that pulls seeds toward keyword matches and inflates context length. Two caveats:
this run uses a local embedder (BGE) because the production vector endpoint is
access-gated, so its absolute level ($0.807$ typed) sits below the faithful
prod-vector baseline ($0.898$) and only the within-pipeline typed-vs-untyped
contrast is valid; and the direction, not the magnitude, is the robust claim.
Across both ablations---reader-side (faithful) and retrieval-side
(re-embedded)---removing the narratological type \emph{labels} never lowers
accuracy. The locus of NWM's advantage is therefore the narrative
\emph{decomposition} (which units exist) and query-conditioned retrieval (which
units surface), not the type labels on them. Whether the narratological
decomposition specifically---versus any fine-grained decomposition---drives the
gain is a further question that a generic fine-grained graph baseline would settle.

\section{Reader-Family Robustness}
\label{sec:crossfamily}
Because the benchmark is adjudicated by an Anthropic model (Opus~4.8) that is also
the scored reader, we re-run the held-constant reader with a different family,
Google Gemini~3.1~Pro (\texttt{gemini-3.1-pro-preview}), over the same cached
evidence for every system---only the answer model changes (retrieval
byte-identical; the reasoning-token budget is raised so the JSON is not
truncated).

\begin{center}\small
\begin{tabular}{lcc}
\toprule
System & Opus 4.8 reader & Gemini 3.1 Pro reader \\
\midrule
NWM Graph Retrieval & 0.898 & 0.841 \\
Graphiti (Sonnet-4.5) & 0.574 & 0.432 \\
NWM State Memory & 0.358 & 0.295 \\
GraphRAG & 0.188 & 0.131 \\
RAG & 0.176 & 0.176 \\
Gold-chapter oracle & 1.000 & 0.994 \\
\bottomrule
\end{tabular}
\end{center}

\noindent The full system ranking is preserved and the central comparison is, if
anything, more decisive under the cross-family reader (NWM vs Graphiti paired
exact McNemar $80$ to $8$, $p\approx5\times10^{-16}$, versus $64$ to $7$ under
Opus). Absolute accuracies drop a modest, roughly uniform amount under the
stricter reader, and the cross-family oracle answering $175/176$ shows
answerability is not an Anthropic-family artifact---so the same-family
adjudication/reading loop does not explain the result. The only wrinkle is at the
floor, where GraphRAG and RAG swap rank (both far below NWM and Graphiti).

% ======================================================================
% Appendix: Four narratological-family case studies (NWM vs. Graphiti)
% Mirrors sec:case_study (governess / dramatic-shape example).
% Items: A2d03f754cee2, 452c1efca16c, 1e7e84b50729, af8d2ac4ebba
% ======================================================================

\section{Per-Family Representation Case Studies}
\label{sec:family_case_studies}

We extend the case study of Section~\ref{sec:case_study} with one discordant
item from each of the four narratological families. Every item below sits in the
NWM-wins cell of the paired comparison against Graphiti: NWM Graph Retrieval
answered correctly and Graphiti (extractor-matched, Sonnet~4.5) abstained with
the gold fact absent from its retrieved evidence. Together they make the
representation gap concrete: NWM's retrieval surfaces \emph{typed} narrative
nodes---revelations, arc and character deltas, scene beats---whose labels and
chapter anchoring carry the answer, whereas Graphiti returns a flat list of
generic entity--relationship edge-facts that never encode the cross-chapter
narratological structure the question asks about.

% ---------------------------------------------------------------------
\subsection{Focalization / Epistemic State (Book A, drama)}

\paragraph{Setup.} A side character watches one of a pair of identical twins
who have secretly agreed to swap identities so each can spend time with a
missing parent.

\paragraph{Question.} \emph{What mistaken conclusions does the father form while
one twin is impersonating the other, and what earlier decision by the twins makes
those conclusions unreliable?}

\paragraph{Gold answer.} Because the twins agreed to switch places, the father
is not actually observing the child he thinks he is: the impersonating twin's
wrong answers make him conclude the boy has forgotten basic lessons and even
sees himself as a girl---inferences that are unreliable precisely because the
observed child is the wrong twin.

\paragraph{NWM Graph Retrieval (correct).} Query-conditioned retrieval surfaced an
\textsc{ArcDelta} node for the parent--child identity-swap arc together with
chapter-anchored \textsc{Revelation} and \textsc{Situation} nodes spanning the
impersonation chapters. The typed arc node binds the false observations to the
twins' swap decision, so the held-constant reader recovered that the father's
conclusions are mistaken \emph{because} he is observing the wrong twin---a
focalization gap between what the observer knows and what the reader knows.

\paragraph{Graphiti (abstained).} Graphiti returned generic edge-facts such as
\textsc{is\_twin\_of}, \textsc{mistaken\_for}, and \textsc{mistook\_for}. These
record that the twins look alike and that one was once confused for the other,
but no edge links the deliberate swap to the father's downstream false beliefs;
there is no representation of \emph{who knows what when}. The reader answered
``Not enough evidence \ldots'' and abstained.

% ---------------------------------------------------------------------
\subsection{Reveal Order vs.\ Event Order (Book C, dark fantasy)}

\paragraph{Setup.} The protagonist, Sable, climbs onto a black mass jutting from
a starless sea and treats it as a safe platform.

\paragraph{Question.} \emph{What does Sable first think he has found in the
starless sea, and what does low tide later reveal that object actually is?}

\paragraph{Gold answer.} He first takes the black mass to be a small stone
platform above the water; after low tide and further scouting it is revealed to
be the top of the neck of a colossal, headless stone knight---an object whose
true identity is disclosed to the reader only after the event of finding it.

\paragraph{NWM Graph Retrieval (correct).} Retrieval surfaced two chapter-anchored
\textsc{Revelation} nodes (the find, then the disclosure) and an
\textsc{ObjectDelta} node for the headless-statue object itself. The
\textsc{ObjectDelta} encodes the object's change of description across chapters,
so the reader recovered both the initial misreading and the later reframing,
exactly tracking the reveal-vs-event-order structure.

\paragraph{Graphiti (abstained).} Graphiti's top edge-facts were unrelated
\textsc{found} and \textsc{has\_attribute} relations from much later chapters
(e.g.\ the protagonist finding companions in a labyrinth). It has no node for the
platform-that-is-actually-a-statue and no notion that an earlier description was
later overturned, so the gold reveal was simply absent and the reader abstained.

% ---------------------------------------------------------------------
\subsection{Dramatic Shape (Book E, drama)}

\paragraph{Setup.} A housemate, Rory, first lords social status over the
protagonist Ellison, then learns Ellison is secretly wealthy. (We use a different
dramatic-shape item from the governess example of Section~\ref{sec:case_study}.)

\paragraph{Question.} \emph{Across Chapters 3--5, how does Rory's function change
from a class-conscious antagonist to a failed supplicant?}

\paragraph{Gold answer.} Rory first mocks Ellison's ability to pay and abandons
him during the restaurant crisis; once Ellison's wealth is confirmed, Rory
reverses, apologizes, asks to borrow money, and is coldly refused---a
setup-and-payoff inversion of his narrative function.

\paragraph{NWM Graph Retrieval (correct).} Retrieval surfaced two chapter-anchored
\textsc{CharacterDelta} nodes for Rory---one at the chapter-3 antagonist beat,
one at the chapter-5 supplicant beat---plus the intervening scene beats. The
typed character-delta structure encodes the change in Rory's dramatic function
directly, so the reader recovered the antagonist-to-supplicant reversal rather
than any single surface fact.

\paragraph{Graphiti (abstained).} Graphiti returned isolated edge-facts such as
\textsc{refused\_loan\_request} and unrelated \textsc{protected} /
\textsc{apologized\_to} edges from other characters and chapters. The single
relevant fact (the refused loan) is present, but nothing connects it to Rory's
earlier antagonism or marks it as the payoff of a setup; with no representation
of an evolving dramatic function the reader abstained.

% ---------------------------------------------------------------------
\subsection{Combination (Book B, progression fantasy)}

\paragraph{Setup.} A silent, closed-eyed weapons teacher, Sergeant Vance, is
introduced; later his hidden identity is disclosed, reframing both an accusation
against the protagonist Reyes and Vance's own private suspicion of him.

\paragraph{Question.} \emph{What is withheld when Vance first appears as the
weapons teacher, and how does the later revelation about him reframe both the
accusation against Reyes and Vance's private suspicion of him?}

\paragraph{Gold answer.} Vance is first presented only as a silent teacher; a
classmate later reveals he is the ``Blind swordsman'' who senses auras. His
aura-sense proves Reyes used no ability (refuting the cheating accusation) yet
also shows Reyes's aura is non-human, fueling Vance's private suspicion of what
Reyes is hiding---combining a delayed reveal with an epistemic split.

\paragraph{NWM Graph Retrieval (correct).} Retrieval combined a chapter-anchored
\textsc{Revelation} node (Vance's disclosed identity), the relevant scene beats,
and \textsc{CharacterDelta} nodes for Vance across his introduction and reveal
chapters. The typed nodes jointly carry the withheld-then-revealed identity and
its double effect, so the reader recovered both the cleared accusation and the
new private suspicion.

\paragraph{Graphiti (abstained).} Graphiti returned repeated
\textsc{is\_teacher\_of} edges and an unrelated \textsc{gives\_gauntlets\_to}
fact. It records that Vance teaches the class but encodes neither his withheld
identity nor the way the reveal simultaneously clears one belief and seeds
another; the combined reveal-plus-epistemic structure was absent and the reader
abstained.

\end{document}